%% file: root_arxiv.tex
\documentclass[letterpaper, 10 pt, journal, twoside]{IEEEtran}

\usepackage{amsmath,amsfonts}
\usepackage{algorithmic}
\usepackage{array}
\usepackage[caption=false,font=normalsize,labelfont=sf,textfont=sf]{subfig}
\usepackage{textcomp}
\usepackage{stfloats}
\usepackage{url}
\usepackage{verbatim}
\usepackage{graphicx}
\def\BibTeX{{\rm B\kern-.05em{\sc i\kern-.025em b}\kern-.08em
    T\kern-.1667em\lower.7ex\hbox{E}\kern-.125emX}}
\usepackage{balance}

\PassOptionsToPackage{hyphens}{url}\usepackage{hyperref}

\hypersetup{
    colorlinks=true,
    linkcolor=blue,
    filecolor=blue,      
    citecolor=blue,
}

\usepackage{math}
\usepackage{tablefootnote}

\newcommand{\spheading}[2][10em]{
  \rotatebox{90}{\parbox{#1}{\centering \small {\textbf{#2}}}}}

\newcolumntype{C}[1]{>{\centering\let\newline\\\arraybackslash\hspace{0pt}}m{#1}}

\newcommand{\eg}{{\em e.g.,~}}
\newcommand{\ie}{{\em i.e.,~}}

\title{VLM-Social-Nav: Socially Aware Robot Navigation \\ through Scoring using Vision-Language Models}

\author{
Daeun Song$^{1}$$^{\star}$, Jing Liang$^{2}$, Amirreza Payandeh$^{1}$, Amir Hossain Raj$^{1}$, Xuesu Xiao$^{1}$, and Dinesh Manocha$^{2}$%

\thanks{$^{\star}$ The majority of the work is conducted as a postdoctoral researcher at the University of Maryland.}
\thanks{$^{1}$D. Song, A. Payandeh, A. H. Raj, and X. Xiao are with the Department of Computer Science, George Mason
University, Fairfax, VA 22030 USA (e-mail: dsong26@gmu.edu; apayande@gmu.edu; araj20@gmu.edu; xiao@gmu.edu).}
\thanks{$^{2}$J. Liang and D. Manocha are with the Department of Computer Science, University of Maryland, College Park, MD 20742 USA (e-mail: jingl@umd.edu; dmanocha@umd.edu).}

}

\begin{document}


\maketitle

\input{v1_arxiv}
\bibliographystyle{IEEEtran}
\bibliography{IEEEabrv,ref}


\end{document}

%% file: v1_arxiv.tex
\begin{abstract}
We propose VLM-Social-Nav, a novel Vision-Language Model (VLM) based navigation approach to compute a robot's motion in human-centered environments. Our goal is to make real-time decisions on robot actions that are socially compliant with human expectations. We utilize a perception model to detect important social entities and prompt a VLM to generate guidance for socially compliant robot behavior. VLM-Social-Nav uses a VLM-based scoring module that computes a cost term that ensures socially appropriate and effective robot actions generated by the underlying planner. Our overall approach reduces reliance on large training datasets and enhances adaptability in decision-making. In practice, it results in improved socially compliant navigation in human-shared environments. We demonstrate and evaluate our system in four different real-world social navigation scenarios with a Turtlebot robot. We observe at least $27.38\%$ improvement in the average success rate and $19.05\%$ improvement in the average collision rate in the four social navigation scenarios. Our user study score shows that VLM-Social-Nav generates the most socially compliant navigation behavior. 

\end{abstract}


\section{Introduction}\label{sec:intro}
Mobile robots integrated into diverse indoor and outdoor human-centric environments are becoming increasingly prevalent. These robots serve various functions, ranging from package and food delivery~\cite{starship} to service~\cite{dilligent} and home assistance~\cite{astro}. Overall, these roles necessitate interaction with humans and navigating seamlessly through public spaces with pedestrians. In such dynamic scenarios, it is important for the robots to engage in socially compliant interactions and navigation~\cite{mirsky2024conflict, mavrogiannis2023core}. 

This paper focuses on the challenges of social navigation~\cite{mavrogiannis2023core}. It addresses the ability of robots to navigate while adhering to social etiquette, especially contextual appropriateness, which requires robots to understand environment contexts, current tasks, and interpersonal behaviors. Therefore, navigating socially across varying contexts presents distinct challenges~\cite{francis2023principles, mavrogiannis2023core, mirsky2024conflict}, including ensuring safety, comfort, and politeness, as well as adhering to social norms.

\textit{Inferring contextually appropriate navigation behaviors is challenging.} Humans have various behaviors and the environmental or task contexts cannot be easily categorized~\cite{mavrogiannis2023core}. 
A common strategy to handle the challenge is by learning-based approaches to learn the complicated contexts empirically. Imitation Learning (IL) is a recent emerging paradigm for desired navigation behavior~\cite{hirose2023sacson, raj2024targeted}. This approach enables autonomous robots to navigate socially by learning from human demonstrations. Other learning approaches, such as reinforcement learning have also been used to address this problem~\cite{kretzschmar2016socially}. While both methods demonstrate promising results in real-world settings, substantial datasets~\cite{thorDataset2019, karnan2022scand, nguyenmusohu} for training and reward engineering are required for their successful application and it is hard to generalize. 

\begin{figure}[t]
\centering
\includegraphics[width=\linewidth]{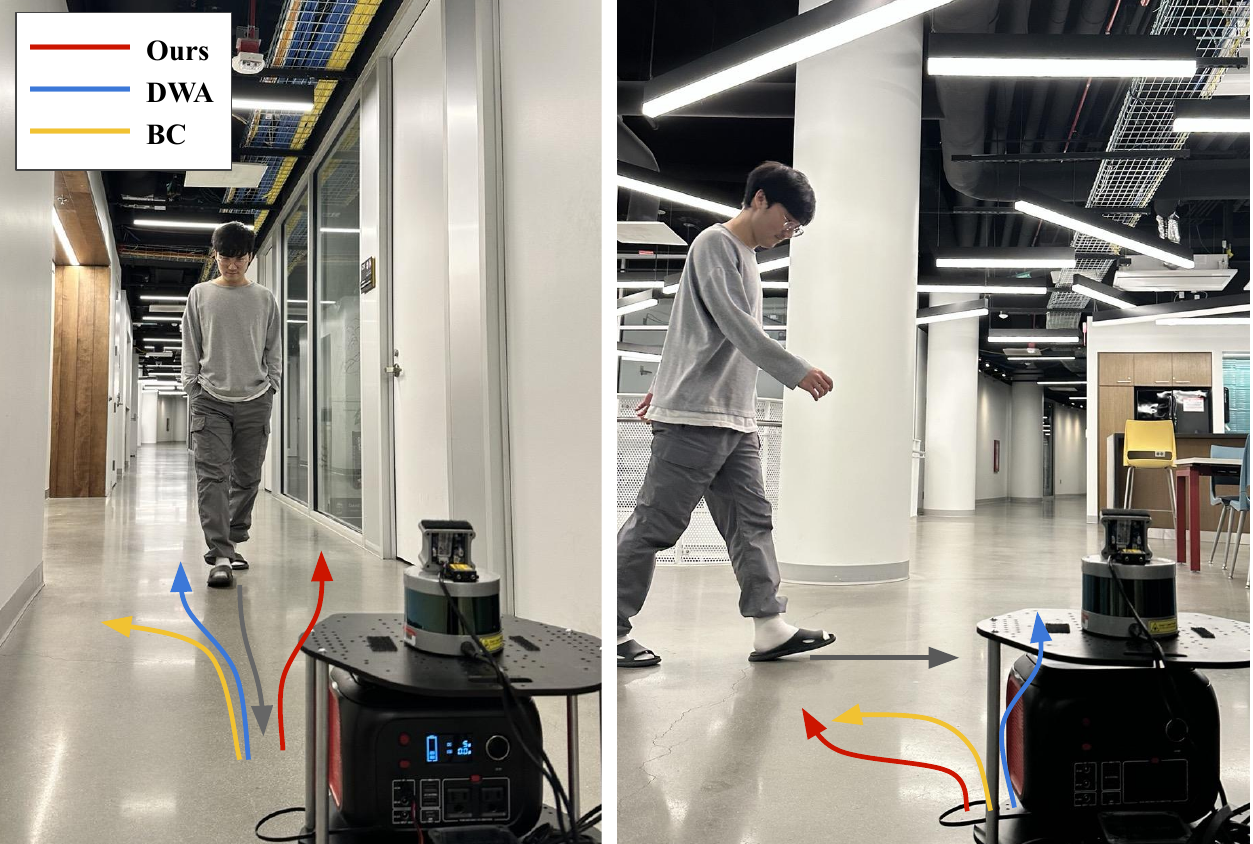} \vspace{-1.0em}
\caption{The trajectories of VLM-Social-Nav (red), DWA (blue), and BC (yellow) approaches in the frontal encountering scenario (left) and the intersection scenario (right). The resulting trajectories show that VLM-Social-Nav demonstrates more socially compliant behavior because it is instructed by a prompt. 
} 
\label{fig:cover}
\end{figure} 

\textit{Language models are inherently well-suited for contextual understanding but not well applied in social navigation.} Recent Large Language Models (LLMs) and Vision-Language Models (VLMs) demonstrate a deep understanding of contextual information and have the potential to perform chain-of-thought~\cite{wei2022chain} and common sense reasoning~\cite{gpt4v2023}. Those processes are inherent to social navigation, especially the challenges of contextual appropriateness and politeness, which require understanding the task/environmental context and the behavior of humans. This capability has also been evaluated across diverse domains of robotics, including human-like driving scenarios~\cite{wen2024road} and autonomous robot navigation~\cite{shah2022lmnav}. However, using language models for social navigation is not well explored, the language models suffer from high latency for real-time navigation, and the issue impedes the smoothness and efficiency of human-robot social interaction. 

{\bf Main Results: } In this paper, we present VLM-Social-Nav, a new approach that uses VLMs to interpret contextual information from robot observation to help autonomous robots improve their navigation abilities in human-centered environments. We leverage a VLM to analyze and reason about the current social interaction and generate an immediate \textit{preferred robot action} to guide an underlying motion planner. {We formalize the concept of \textit{social cost} and the problem definition of social robot navigation suitable for language descriptions. The social cost is defined as how well a robot's behavior aligns with socially acceptable norms, \ie the behavior a human would likely exhibit.} Our VLM-based scoring module computes the social cost, which is used for a
bottom-level motion planner to output appropriate robot actions. To overcome the limitation of existing VLMs' latency issue, we utilize a state-of-the-art perception model (\ie YOLO~\cite{YOLO2023}) to detect key entities that are used for social interactions (\eg humans, gestures, and doors) and query a VLM to generate socially compliant navigation behavior and compute the social cost. We demonstrate VLM-Social-Nav in four different indoor scenarios with human interactions. Unlike previous social navigation approaches, VLM-Social-Nav can better navigate through social scenarios by interpreting the situation based on \textit{common sense} without any dedicated training on a large dataset. Some of our main results include:
\begin{itemize}
  \item We propose VLM-Social-Nav, a novel approach for social robot navigation, by integrating VLMs with optimization-based or scoring-based motion planners and a state-of-the-art perception model for better VLM efficiency. 
  \item We propose a VLM-based scoring module that translates the current robot observation and textual instructions into a relevant social cost term. This cost term is used for the bottom-level motion planner to output appropriate robot action.
  \item We evaluate VLM-Social-Nav in four different real-world indoor social navigation scenarios 
  along with a user study and compare the results with a Dynamic-Window Approach (DWA)~\cite{fox1997dynamic} and Behavior Cloning (BC)~\cite{pomerleau1988alvinn} method trained on a state-of-the-art large Socially CompliAnt Navigation Dataset (SCAND)~\cite{karnan2022scand}. VLM-Social-Nav achieves at least $27.38\%$ improvement in average success rate and $19.05\%$ improvement in average collision rate in four scenarios. The user study score shows that VLM-Social-Nav generates the most socially compliant navigation behavior. 
  
\end{itemize}

{More related works and discussions can be found in the technical report} of our manuscript~\cite{song2024socially}.

\section{Related Work}
\label{sec:prev}
In this section, we give an overview of existing works related to safety requirements and different challenges of contextual appropriateness in social robot navigation, and Large Foundation Models (LFMs) for robot navigation. 

\subsection{Safety Requirement of Social Navigation}
For social navigation, safety is a basic requirement for interacting with humans and navigating dynamic scenarios~\cite{liang2021vo, liang2021crowd}. DWA~\cite{fox1997dynamic} is a well-known collision-free navigation method that calculates collision constraints and selects the best feasible action. 
Velocity-Obstacle (VO)-based approaches are more efficient and can be used to simulate the actions of crowds~\cite{best2014densesense}, but they do not take uncertainties into account. PRVO~\cite{gopalakrishnan2017prvo} and OFVO~\cite{liang2021vo} handle the perception uncertainties, but those approaches require a hard threshold for planning. To deal with this issue, learning-based methods empirically train the policies by demonstrations~\cite{sun2021motion} or use reinforcement learning to train the robot in a simulator and implement it in real-world scenarios~\cite{liang2021crowd, aradi2020survey}. However, learning-based approaches require a significant amount of data or realistic simulators to learn the task. Social interactions are highly nuanced and context-dependent. Simulating these interactions accurately requires sophisticated models of human behavior and interaction dynamics, which is not trivial.

\subsection{Contextual Appropriateness of Social Navigation}
Researchers have developed various methods to incorporate social awareness into mobile robot navigation. Creating such systems is complex, requiring advanced perception and reasoning to navigate environments shared with humans and robots \cite{mirsky2024conflict}. Defining social navigation varies across cultures and platforms. Assessing social compliance, beyond safety, depends on the scenario and requires contextual consideration. Various methodologies are employed to address this challenge, with a significant focus on enhancing learning methods through reinforcement learning, learning from demonstration (particularly by analyzing examples of human trajectories or robots operated by humans), and the utilization of simulated datasets \cite{7539621, li2018role, nazeri2024vanp}. 
Additionally, various datasets have been collected for this purpose. 
SCAND~\cite{karnan2022scand} and MuSoHu~\cite{nguyenmusohu} are two recent large-scale social human navigation datasets in many natural human-inhabited public spaces for robots to learn similar, human-like, socially compliant navigation behaviors. 
Although extensive research has explored various machine learning techniques, Vision-Language Models (VLMs) have not yet been applied to the social navigation problem, despite their strong potential for contextual analysis. 

\begin{figure*}[t]
\centering
\includegraphics[width=0.8\linewidth]{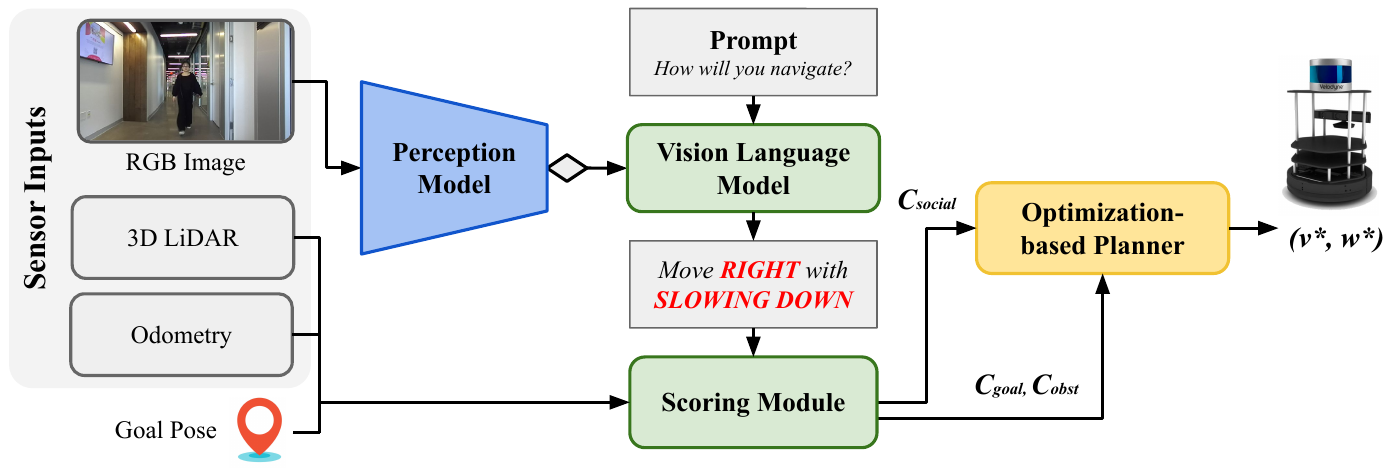}\vspace{-0.5em}
\caption{The overall system architecture of VLM-Social-Nav. Our real-world perception (vision) model detects important social entities (\eg humans, gestures, and doors) in real time and prompts the VLM-based scoring module to compute social cost $\cc_\textrm{social}$, which is used to generate socially compliant robot action. 
} \vspace{-0.5em}
\label{fig:overview}
\end{figure*} 

\subsection{Large Foundation Models for Navigation}
Recent advancements in Language Foundation Models (LFMs)~\cite{bommasani2022opportunities}, encompassing VLMs and LLMs, show significant potential in robotic navigation. 
SayCan~\cite{Ahn2022} integrates LLMs for high-level task planning. GPT-Driver~\cite{ mao2023gptdriver} evaluates the performance of GPT-3.5 in simulation for autonomous driving, framing motion planning as a language modeling problem. L3MVN~\cite{yu2023l3mvn} constructs semantic maps of environments and utilizes LLMs to reach long-term goals, while LLaDA~\cite{li2024driving} enables autonomous vehicles to adapt to diverse traffic rules across regions. LM-Nav~\cite{shah2022lmnav} utilizes GPT-3 and CLIP~\cite{radford2021learning} to navigate outdoor environments based on natural language instructions, combining language and visual cues for optimal path planning. 
Despite their powerful capabilities in contextual understanding and commonsense reasoning, language models have not been extensively investigated for social navigation. 
Our approach proposes a novel method to navigate robots in a socially compliant manner.

\section{Approach}\label{sec:ap}

In this section, we define the social navigation problem and describe VLM-Social-Nav in detail. 

\subsection{Problem Definition}

Navigation is the task of generating and following an efficient collision-free path from an initial location to a goal~\cite{mavrogiannis2023core}. In general, the overall system consists of a global planner and a local planner. A global planner is designed to find a collision-free path to reach a goal, while a local planner aims to navigate the robot through its immediate surroundings, making real-time adjustments to deal with vehicle dynamics and surrounding obstacles. 

For social robot navigation, humans are no longer perceived only as dynamic obstacles but also as social entities~\cite{mirsky2024conflict}. It necessitates integrating social norms into robot behaviors. 
We define the social robot navigation problem as a {\em Markov Decision Process (MDP)}: $\langle \cs, \ca, \ct, \cc \rangle$, where $\s = (x, y, \theta) \in \cs$ is a state consisting of a robot pose, $\a = (v, w) \in \ca$ is an action consisting of a linear and an angular velocity of the robot, $\ct: \cs \times \ca \rightarrow \cs$ is the transition function characterizing the dynamics of the robot, and $\cc: \cs \times \ca \rightarrow \RR $ is a cost function. Given a cost function $\cc$, the motion planner finds $(v^*, w^*)$ that minimizes the expected cost. The cost function takes the following form:
\begin{equation}\label{eq:cost}
\begin{aligned}
    \cc(\s,\a)=\alpha\cdot \cc_\textrm{goal}+\beta\cdot \cc_\textrm{obst}+\gamma\cdot \cc_\textrm{social},
\end{aligned}
\end{equation}
where $\cc_\textrm{goal}$ encourages movement toward the goal, $\cc_\textrm{obst}$ discourages collisions with obstacles, and $\cc_\textrm{social}$ encourages the robot to follow the social norms. $\alpha$, $\beta$, and $\gamma$ are non-negative weights for each cost term. 

The social cost term $\cc_\textrm{social}$ encompasses various factors that govern human-robot interactions in shared environments. Defining them mathematically poses challenges. For VLM-Social-Nav, we define $\cc_\textrm{social}$ as:
\begin{equation}\label{eq:social}
\begin{aligned}
    \cc_\textrm{social}= \|\cb - \cb_{h} \|,
\end{aligned}
\end{equation}
where $\cb$ is a navigation behavior and $\cb_{h}$ is a navigation behavior humans would adopt in accordance with social conventions. Minimizing the deviation between them will encourage the robot to emulate socially acceptable human behaviors. While $\cb_{h}$ can be obtained through various methods, including large datasets~\cite{thorDataset2019, karnan2022scand, nguyenmusohu}, we leverage the power of a VLM to compute appropriate behavior based on its rich contextual understanding and nuanced interpretations from perceived images and given prompts. We elaborate further in Section~\ref{sec:vlm}.

\subsection{VLM-based Social Navigation Architecture}

Fig.~\ref{fig:overview} highlights the overview of VLM-Social-Nav. Our approach is based on an autonomous navigation system that integrates a perception layer with an optimization-based motion planner. The motion planner processes sensor inputs and generates a robot action that minimizes the cost function $C$. 

While LiDAR detects geometric information useful for obstacle avoidance, RGB images provide contextual details of the current environment. They contain rich information crucial for social navigation. 
To enhance navigation capabilities within social contexts, we propose a VLM-based scoring module. VLMs excel in contextual understanding, interpreting scenes not solely based on visual features but also considering social dynamics{\cite{duan2024manipulate}}. VLMs generate socially appropriate robot actions based on current observations and input instructions. Our VLM-based scoring module then calculates a cost term to be used by the motion planner. 

While VLMs can generate navigation behaviors that comply with social norms, continuously querying large VLMs for new responses is prohibitively computationally expensive for real-time navigation. To address this challenge, we incorporate a real-time perception model. This model identifies social entities such as humans, gestures, and doors as the robot navigates its environment. Our VLM-based scoring module activates only when significant social cues are detected, ensuring that the social cost term is integrated only when necessary, \ie when there is any human interaction involved. This approach reduces the VLM queries and facilitates real-time navigation efficiency for our approach. Algorithm~\ref{alg:vsn} summarizes an overview of our VLM-Social-Nav process.

\begin{algorithm}[tb]
    \caption{VLM-based Social Navigation}\label{alg:vsn}
    \SetKwInOut{Input}{Input}
    \SetKwInOut{Output}{Output}
    \Input{RGB image $\ci$, LiDAR point cloud $\cl$, prompt $\cp$, goal position $\p_g$}
    Initialize robot position $\p_r$\;
    \While{not at goal position $\p_r \neq \p_g$}{
      $\ci \gets$ Read image sensor data\;
      $\cl \gets$ Read LiDAR sensor data\;
      $e \gets \text{Perception Model}(\ci)$\;
      \ForAll{possible actions $\a$}{
        $\cc_\textrm{social} \gets 0$\;
        \uIf{social entities detected in $e$}{
            $\cb_h = \text{VLM}(\ci, \cp ,\a)$\;
            $\cc_\textrm{social} = \text{VLM-based scoring}(\cb_h)$\;
        }
        Calculate the total coast $\cc = \alpha\cdot \cc_\textrm{goal}+\beta\cdot \cc_\textrm{obst}+\gamma\cdot \cc_\textrm{social}$\;
      }
      Find action $\a$ with minimal cost $\cc$ and execute\;
      Update robot position $\p_r$\;
    } 
\end{algorithm} 

\subsection{VLM-based Scoring Module} \label{sec:vlm}

VLM plays a crucial role in VLM-Social-Nav in inferring immediate socially compatible navigation behavior $\cb_h^{t+1}$ based on its pre-trained large internet-scale dataset:
\begin{equation}\label{eq:vlm}
\begin{aligned}
    \cb_h^{t+1} = \text{VLM}(\ci^t, \cp ,\a^t) ,
\end{aligned}
\end{equation}
where $\ci^t$ is an RGB image from the robot view at time $t$, $\cp$ is a textual prompt, and $\a^t$ is a current robot action at time $t$. Inspired by In-Context Learning (ICL), our prompt $\cp$ is designed to leverage the VLM's reasoning abilities through zero-shot examples. This approach offers an interpretable interface, mirroring human reasoning and decision-making processes, without extensive training~\cite{min2022rethinking}.

Our VLM-based scoring module starts from the insight that the action space of a mobile robot can be readily mapped to linguistic terms. For example, the action ``move forward at a constant speed'' can be linked to a linear velocity of $v^{t}$ m/s and an angular velocity of 0. 
The heading direction on the left indicates a positive value of $w^{t}$, while the direction on the right indicates a negative value. Leveraging this understanding, we structure the output of the VLM into a linguistic format comprising the heading and the speed. Subsequently, our scoring module extracts $\cb_h^{t+1} \mapsto (v_h^{t+1}, w_h^{t+1}) \in \ca$ from these tokens; $v_h^{t+1} = v^t + \delta_{s}$, where $\delta_{s}$ is derived from the response for the speed; $w_h^{t+1} = \delta_{d}$, where $\delta_{d}$ is derived from the response for the heading. 
Thus, the social cost term for the next time step can be calculated:
\begin{equation}\label{eq:social_vlm}
\begin{aligned}
    \cc_\textrm{social}^{t+1}= w_{l} \cdot \| v - v_h^{t+1} \| + w_{a} \cdot \| w - w_h^{t+1} \|,
\end{aligned}
\end{equation}
where $w_{l}$ and $w_{a}$ are non-negative weights. Given all the cost terms, our low-level optimization-based motion planner finds the robot action $(v^*, w^*)$ that minimizes the cost. 

\begin{figure}[tb]
\begin{GrayBox}      
    \footnotesize{\textbf{Input Image:} \vspace{0.5em} \\ }
    \includegraphics[width=0.55\linewidth]{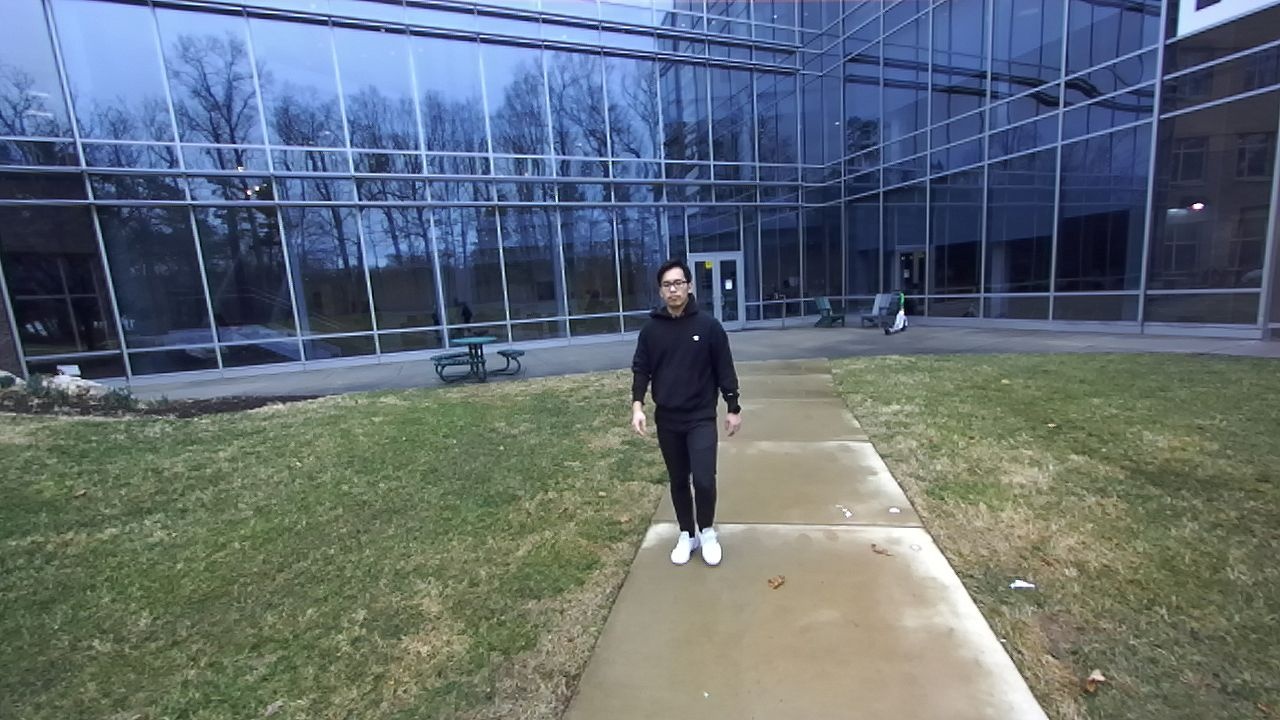} \\ \\
    \footnotesize{\textbf{Input Prompt:} \vspace{0.5em}} \\
    \footnotesize{
    \textit{Task:} \\
    How will you navigate concerning the person in your view? You will need to follow general walking etiquette. \vspace{0.5em} 
    
    \textit{Ego state:} \\
    - heading direction: \textcolor{blue}{straight}\\
    - linear velocity: \textcolor{blue}{0.28} \vspace{0.5em}

    \textit{Remember:} \\
    - Move to the right when passing by a person.\\
    - Do not obstruct others' paths.\\
    - ... \vspace{0.6em}

    \textit{Answer Format:} \\
    Move \textcolor{red}{DIRECTION} with \textcolor{red}{SPEED} \\
    - options for DIRECTION: left, straight, right \\
    - options for SPEED: slow down, speed up, constant, stop
    }
\end{GrayBox} \vspace{-0.5em}
\caption{An example input image ($\ci^t$) and prompt ($\cp$) used in VLM-Social-Nav. Parameterized inputs ($\a^t$) are highlighted in blue. Formatted outputs specifying the heading ($\delta_d$) and the speed ($\delta_s$) are highlighted in red. The example input data is one of the frontal approach scenarios from MuSoHu~\cite{nguyenmusohu}. 
} \vspace{-1.0em}
\label{fig:prompt}
\end{figure}

Fig.~\ref{fig:prompt} shows an example prompt $\cp$ used in our experiment. We provide a high-level task description along with an image $\ci^t$ captured from the robot's perspective. Furthermore, the current robot action $\a^t=(v^t,w^t) \in \ca$ is provided. {The angular velocity is mapped into corresponding directional instructions based on predefined categories (\ie positive values correspond to \emph{left}, values near zero to \emph{straight}, and negative values to \emph{right}).} 
Supplementary instructions regarding walking etiquette are included. Although the VLM demonstrates proficient navigation abilities in the absence of explicit instructions, offering reasoning guidelines enhances its decision-making processes~\cite{min2022rethinking}. These guidelines not only facilitate comprehensive reasoning and judgment within the VLM but also enable the robot to adapt to specific rules more effectively. For example, in a country where it's customary to walk on the left, we can rephrase the prompt as ``Move to the left when passing by another person.''

\begin{figure*}[htb]
\centering
\begin{tabular}{rc}
\spheading{(a) Frontal Approach} \hspace{-0.6em} & \includegraphics[width=0.9\linewidth]{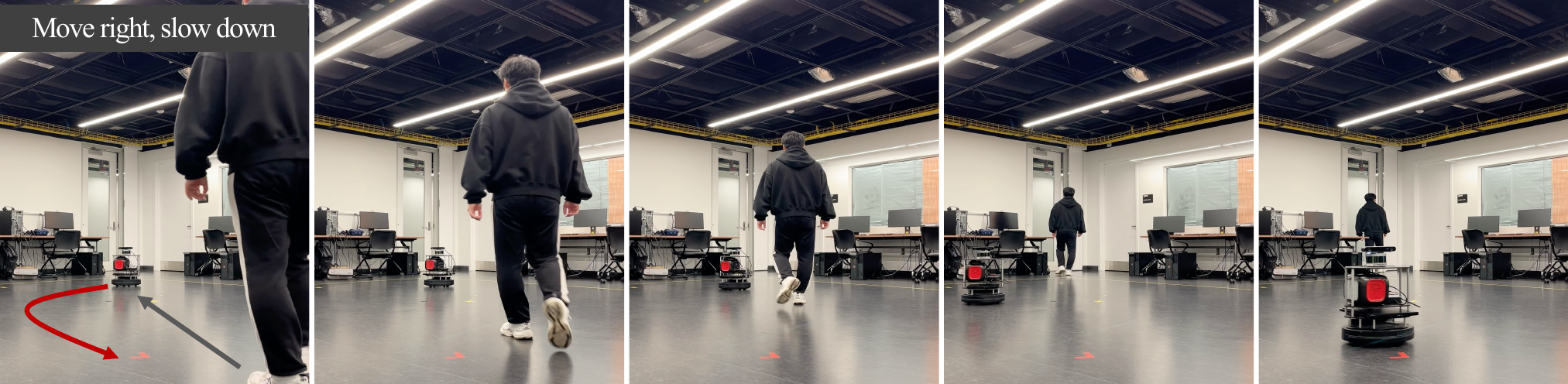} \\
\spheading{(b) Frontal Approach with Gesture}  \hspace{-0.6em} & \includegraphics[width=0.9\linewidth]{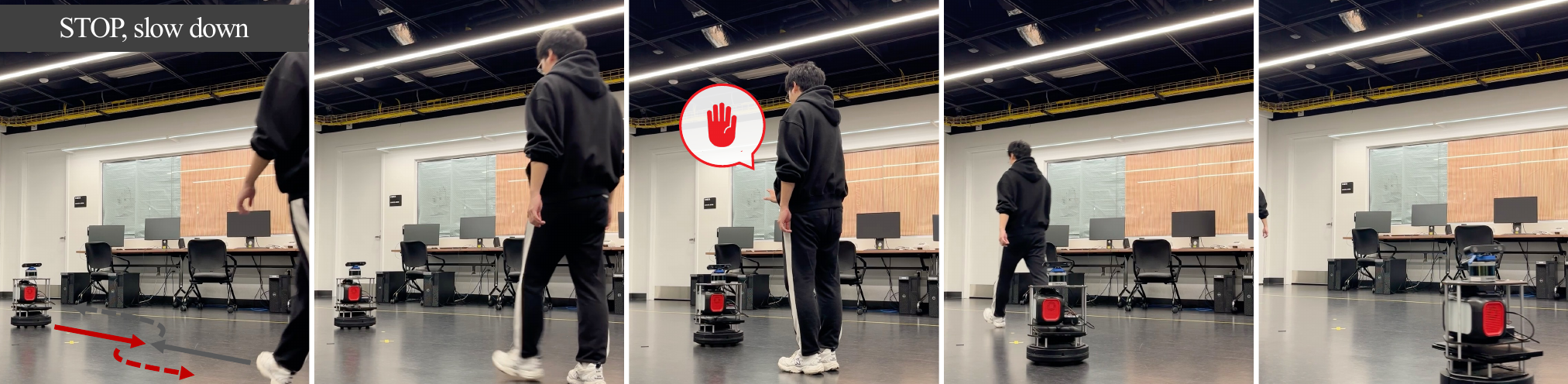} \\
\spheading{(c) Intersection} \hspace{-0.6em}  &  \includegraphics[width=0.9\linewidth]{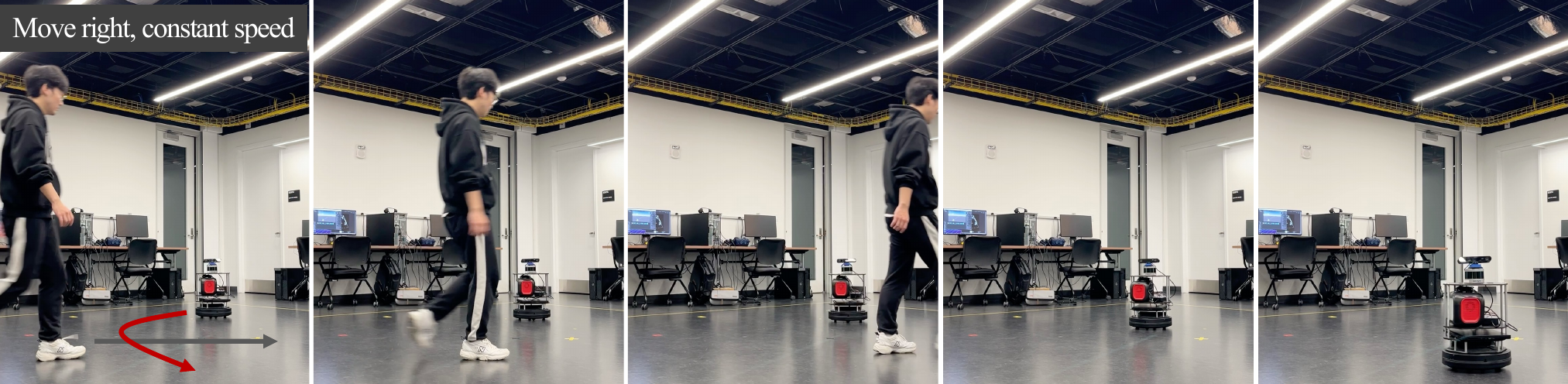} \\
\spheading{(d) Narrow Doorway} \hspace{-0.6em} &  \includegraphics[width=0.9\linewidth]{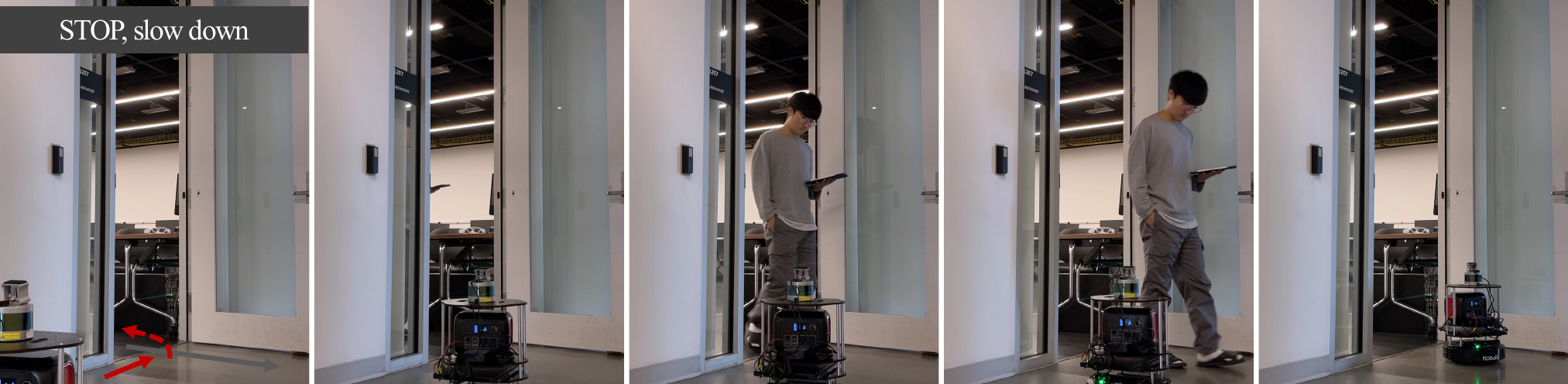} \\
\end{tabular} 
\caption{Qualitative Results: the robot navigation behaviors with VLM-Social-Nav for four social navigation scenarios: (a) Frontal Approach, (b) Frontal Approach with Gesture, (c) Intersection, and (d) Narrow Doorway. The solid gray arrow shows the participant's path. The solid red arrow shows the robot's path. The red and gray dashed arrows show the robot's and participant's paths respectively, after a stop motion. A caption on the top left shows the result from the VLM.}  \vspace{-1.0em}
\label{fig:qualitative}
\end{figure*}

\section{Experiments}\label{sec:exp}

In this section, we describe the details of the implementation and qualitative and quantitative experimental results. 

\subsection{Implementation Details}

VLM-Social-Nav is tested on a Turtlebot 2 equipped with a Velodyne VLP16 LiDAR, a Zed 2i camera, and a laptop with an Intel i7 CPU and an Nvidia GeForce RTX 2080 GPU. 
We use YOLO~\cite{YOLO2023} as our real-world perception model to detect key objects. {In our experiment, we focus on key social cues, \ie humans, doors, and gestures, which are critical considerations when navigating in socially rich environments~\cite{kruse2013human}.} 
Generative Pre-trained Transformer 4 with Vision (GPT-4V)~\cite{gpt4v2023} is used as our VLM to comprehend the social dynamics and output the immediate preferred robot action. {This follows a preliminary study that compared its performance with other large and small VLMs. GPT-4V was able to produce reliable results with high consistency, achieving a reasonable average inference time of around three seconds.} 
We combined our approach with a low-level motion planner DWA~\cite{fox1997dynamic}. 
We compare VLM-Social-Nav with DWA without social cost $\cc_\textrm{social}$ and BC~\cite{pomerleau1988alvinn} trained on a state-of-the-art, large-scale social navigation dataset, SCAND~\cite{karnan2022scand}. The dataset contains various examples of socially compliant navigation behaviors teleoperated by humans including sticking to the right of the road and waiting for a human to pass. {We expect that the model can learn to output socially compliant navigation behaviors like the human demonstrations.}


{Evaluating the social aspects of social robot navigation is inherently challenging~\cite{tsoi2024robot}. To validate VLM-Social-Nav, we carefully follow the social robot navigation studies~\cite{francis2023principles, pirk2022protocol}, which set up the benchmark scenarios and the metrics for measuring social compliance.} 
We present qualitative, quantitative, and user study results in four different social navigation scenarios:
\begin{itemize}
  \item \textbf{Frontal Approach:} A robot and a human approach each other from two ends of a straight trajectory.  
  \item \textbf{Frontal Approach with Gesture:} A robot and a human approach each other from two ends of a straight trajectory. The human recognizes the robot and then gestures for it to stop.
  \item \textbf{Intersection:} A robot and a human cross each other on perpendicular trajectories. 
  \item \textbf{Narrow Doorway:} A robot and a human cross each other's paths by moving through a narrow doorway. 
\end{itemize}

\begin{table*}[tb] 
\centering
\caption{Quantitative Results: performance comparisons using BC~\cite{pomerleau1988alvinn}, DWA~\cite{fox1997dynamic}, and VLM-Social-Nav }
\begin{tabular}{rccccc} 
 \midrule
 \multirow{2}{*}{\textbf{Metric}} & \multirow{2}{*}{\textbf{Method}} & \multicolumn{4}{c}{\textbf{Scenario}} \\
 \cmidrule(lr{0.5em}){3-6}
 & &  (a) Frontal Approach & (b) Frontal Approach  w/ Gesture & (c) Intersection & (d) Narrow Doorway \\
 \midrule
 \multirow{3}{*}{Success Rate (\%) {$\uparrow$}} & BC & 38.10 & 0 & 33.33 & 42.86\\
 & DWA & \textbf{100} & 0 & 90.48 & \textbf{100}\\
 & VLM-Social-Nav & \textbf{100} & \textbf{100} & \textbf{100} & \textbf{100}\\
 \midrule
 \multirow{3}{*}{Collision Rate (\%) {$\downarrow$}} & BC & 42.86 & 66.67 & 28.57 & 38.10\\
 & DWA & 28.57 & 19.05 & 19.05 & 38.10 \\
 & VLM-Social-Nav & \textbf{14.29} & \textbf{0} & \textbf{4.76} & \textbf{9.52} \\
 \midrule
 \multirow{3}{*}{User Study Score {$\uparrow$}} & BC & 2.80 $\pm$ 1.45 & 2.23 $\pm$ 1.54 & 2.80 $\pm$ 1.40 & 2.60 $\pm$ 1.33\\
 & DWA & 3.99 $\pm$ 0.80 & 3.38 $\pm$ 0.64 & 3.57 $\pm$ 0.62 & 3.59 $\pm$ 0.83\\
 & VLM-Social-Nav & \textbf{4.31 $\pm$ 0.72} & \textbf{4.28 $\pm$ 0.56} & \textbf{4.35 $\pm$ 0.70} & \textbf{4.04 $\pm$ 0.74}\\
 \midrule
\end{tabular} \vspace{-0.5em}
\label{table:quantitative}
\end{table*}  

\subsection{Qualitative Result}

{Based on the protocols and principles set by other studies~\cite{francis2023principles, pirk2022protocol}, the robot is expected to behave in a socially compliant way as follows}: 
\begin{itemize}
  \item \textbf{Frontal Approach:} The robot is expected to yield or slow down and modify its original trajectory so that it does not obstruct the human path. Like driving rules in North America, it is conventional to keep on the right. 
  \item \textbf{Frontal Approach with Gesture:} The robot is expected to yield by interpreting the human gesture. 
  \item \textbf{Intersection:} The robot is expected to drive slowly when it approaches the human. It may come to a complete stop or modify its original trajectory to go behind the human to not obstruct the path. 
  \item \textbf{Narrow Doorway:} The robot is expected to wait outside the door and yield to the human. 
\end{itemize}

Fig.~\ref{fig:qualitative} shows snapshots of the resulting robot motion using VLM-Social-Nav in four different scenarios. We demonstrate that VLM-Social-Nav follows the social convention and navigates toward its goal as expected. 
Fig.~\ref{fig:cover} illustrates the resulting trajectories of VLM-Social-Nav in comparison to those of DWA and BC methods. A notable observation is that, while DWA also effectively avoids collisions with individuals, VLM-Social-Nav generates trajectories that align more closely with social norms. For instance, in the frontal approach scenario, while DWA tends to maneuver around the person either to the right or left, VLM-Social-Nav predominantly bypasses the person on the right side.  Similarly, in the intersection scenario, whereas DWA occasionally obstructs the person's path by veering to avoid collision directly in front, VLM-Social-Nav {adjusts its trajectory to pass behind the individual, adapting effectively to the human's movement direction.} Additionally, BC avoids humans but fails to recover and follow the original path. This leads to many failures in reaching the goal. 
The accompanying supplementary video shows the resulting robot motions.

\begin{table}[htb] 
\centering
\caption{Social Compliance Questionnaire}
\begin{tabularx}{\linewidth}{l|l} 
 \midrule
 \multicolumn{2}{l} {\textbf{Scenario 1: Frontal Approach}} \\
 \midrule
 1 & The robot moved to avoid me.\\
 2 & The robot obstructed my path.$^{*}$\\
 3 & The robot maintained a safe and comfortable distance at all times.\\
 4 & The robot nearly collided with me.$^{*}$\\
 5 & It was clear what the robot wanted to do.\\
 \midrule
  \multicolumn{2}{l} {\textbf{Scenario 2: Frontal Approach with Gesture}} \\ 
 \midrule
 6 & The robot maintained a safe and comfortable distance at all times.\\
 7 & The robot slowed down and stopped.\\
 8 & The robot followed my command\\
 9 & I felt like the robot paid attention to what I was doing.\\
 \midrule
 \multicolumn{2}{l} {\textbf{Scenario 3: Intersection}} \\ 
 \midrule
 10 & The robot let me cross the intersection by maintaining a safe and \\ & comfortable distance.\\
 11 & The robot changed course to let me pass.\\
 12 & I felt like the robot paid attention to what I was doing.\\
 13 & The robot slowed down and stopped to let me pass.\\
 \midrule
 \multicolumn{2}{l} {\textbf{Scenario 4: Narrow Doorway}} \\ 
 \midrule
 14 & The robot got in my way.$^{*}$\\
 15 & The robot moved to avoid me.\\
 16 & The robot made room for me to enter or exit.\\
 17 & It was clear what the robot wanted to do.\\
 \midrule
\end{tabularx} \vspace{-1.5em}
\label{table:userstudy} 
\end{table}

\subsection{Quantitative Result}

To further validate VLM-Social-Nav, we evaluate the methods using three different 
metrics. The success rate describes whether the robot reaches the goal. For the frontal approach with gesture scenario, we mark it as successful when the robot reacts to the gesture. The collision rate describes whether the robot collided with the human or other objects in the environment. We also mark it as in collision when we manually intervene to avoid an imminent collision with the human subject or surroundings. The user study score is an average score we obtained from the user study detailed in Section~\ref{sec:userstudy}.

Table~\ref{table:quantitative} reports the results averaged over 21 runs for each method and scenario. The results demonstrate that VLM-Social-Nav, DWA with social cost, outperforms other methods in every metric. DWA excels at following a path smoothly, yet it faces challenges in collision avoidance as it relies solely on the LiDAR sensor and does not consider social compliance. Most of the collisions occurred when DWA navigated in a way that interfered with a person's path, for example, going in front of the person when intersecting.
We also observe that the outcomes of BC varied. At times, when attempting to avoid collisions, it failed to return to its original path and failed to reach the goal. Conversely, there were instances where it didn't attempt collision avoidance at all, resulting in collisions with the participants. For gesture recognition, only our proposed method successfully responded to the participants' gestures. VLM-Social-Nav improves the average success rate by $27.38\%$ and reduces the average collision rate by $19.05\%$ across four social navigation scenarios.


\subsection{User Study}\label{sec:userstudy}

To validate the social compliance of VLM-Social-Nav, we conduct a user study. 
We ask the participants to walk along the predefined trajectory and then to answer questionnaires about the robot motion~\cite{pirk2022protocol} (Table~\ref{table:userstudy}). * denotes negatively formulated questions, for which we reverse-code the ratings to ensure comparability with the positively formulated ones. The three methods are randomly shuffled and repeated three times. Each scenario is tested on seven participants. 
We use a five-level Likert scale to ask participants to rate their agreement with these statements. 

\begin{figure}[tb]
\centering
\includegraphics[width=0.45\linewidth]{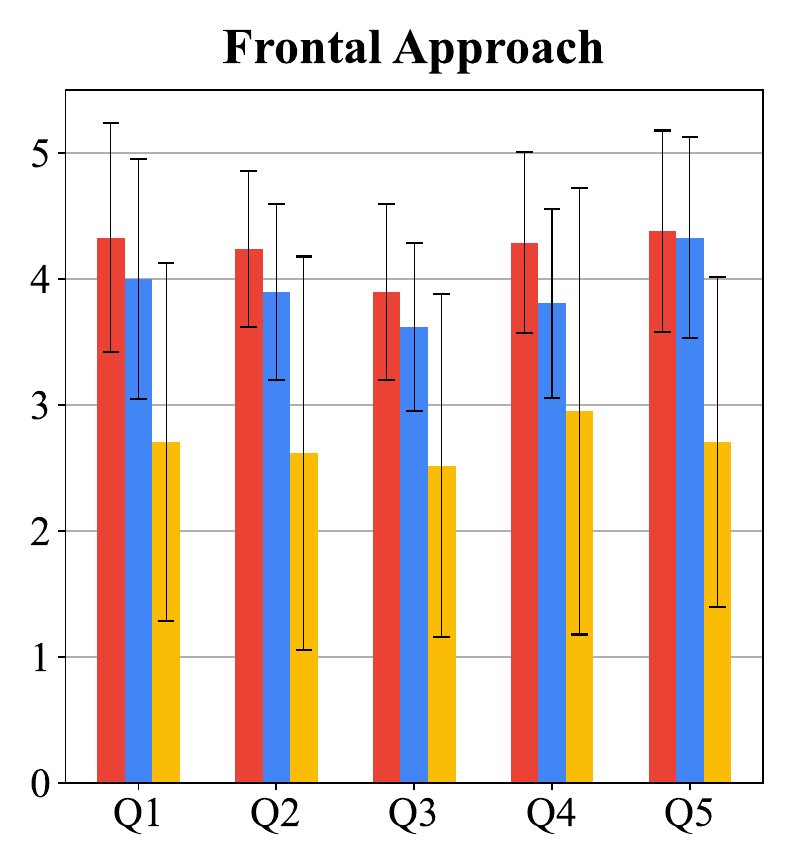}
\includegraphics[width=0.45\linewidth]{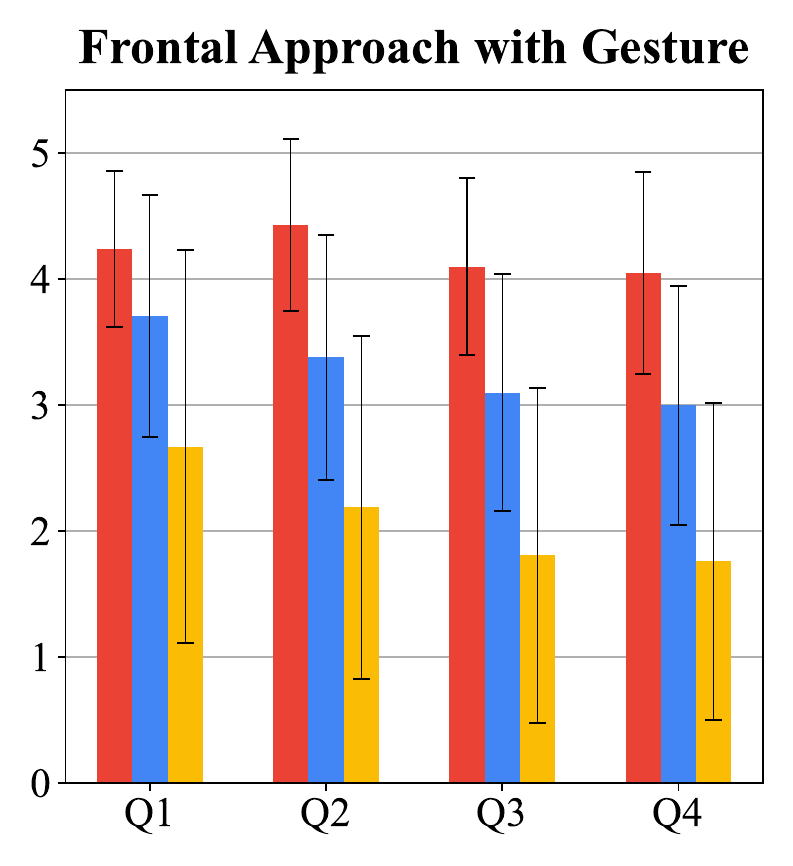} \vspace{-0.5em}
\includegraphics[width=0.45\linewidth]{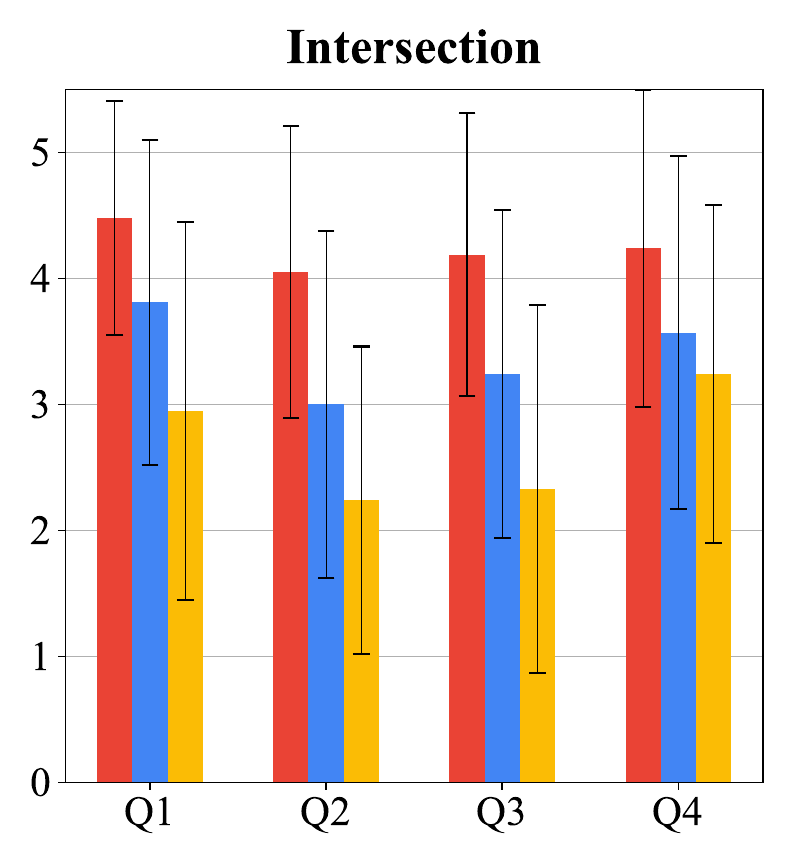}
\includegraphics[width=0.45\linewidth]{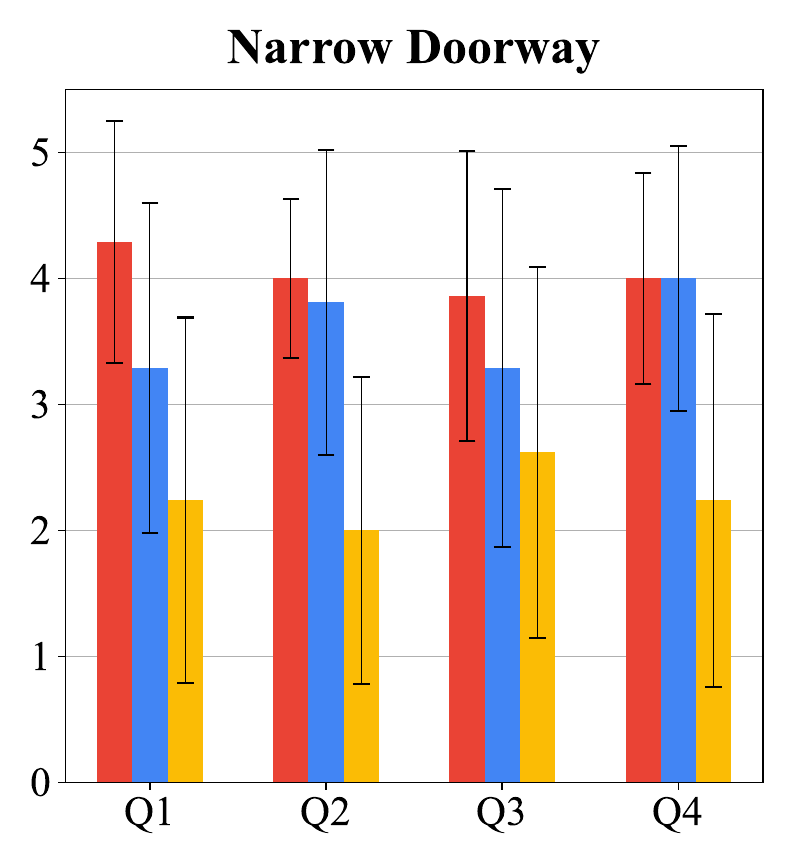}
\includegraphics[width=0.3\linewidth]{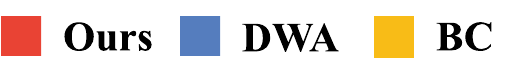} \vspace{-0.5em}
\caption{User Study Average Scores: the per-question average scores for the three methods in each scenario. The results indicate that VLM-Social-Nav earned the highest level of agreement from participants across all questions, highlighting its robust alignment with social norms.} \vspace{-1.0em}
\label{fig:userstudy}
\end{figure}

Fig.~\ref{fig:userstudy} and the user study scores in Table~\ref{table:quantitative} show the study result. The plot shows the per-question average scores for the three methods in each scenario. 
Based on the results, it's evident that VLM-Social-Nav receives the highest level of agreement from participants across all questions, indicating its strong adherence to social norms. The standard error of the BC method was large, indicating that the performance of the BC method was not consistent. The score difference between VLM-Social-Nav and DWA was not large in the narrow doorway scenario. This is because, when attempting to enter the narrow doorway, DWA often failed to find a plan and froze, resembling the result of VLM-Social-Nav, a complete stop in front of the doorway. 

\subsection{Discussion} \label{sec:discussion}

\textbf{Real-time navigation with VLM: }
GPT-4V and similar large VLMs require several seconds to respond to prompts, making continuous querying impractical for real-time navigation tasks. To address this, we optimized VLM-Social-Nav in two ways: first, by formatting prompts and providing predefined choices, which resulted in reduced response times. Second, we minimize queries by using a perception model to detect social cues, allowing for timely VLM queries only when necessary. These choices enable and allow average response times of 2-3 seconds, sufficient for human interaction and navigation. While such a limitation can be problematic in more dynamic scenarios that require frequent interactions, future advancements in fast large language models promise further extensions of our approach.

\textbf{Socially aware navigation with VLM: } 
We observe that VLMs can analyze and reason about social interactions from single images. Using various single images, including those collected by ourselves and from social robot navigation datasets~\cite{nguyenmusohu,karnan2022scand}, VLMs accurately describe scenes and suggest socially compliant navigation strategies with reasons. For instance, for the image shown in Fig.~\ref{fig:prompt}, GPT-4V describes the scene as \textit{a person is walking towards the camera along a sidewalk}. To navigate this situation, GPT-4V advises the robot to \textit{yield the right of way} because \textit{it is generally customary to keep to the right side of the path when encountering someone coming from the opposite direction, similar to driving rules}.
However, despite their powerful capabilities, VLMs can still make mistakes. Therefore, relying solely on VLMs for navigation decisions is not safe. Instead, we incorporate their output as a cost term in our overall decision-making process.


\textbf{More challenging scenarios: } 
Although our robot experiments were conducted only indoors, according to the example in Fig.~\ref{fig:prompt}, VLM-Social-Nav can be extended to outdoor scenarios in more complex environments. VLMs successfully retrieve significant environmental information for outdoor social robot navigation, such as sidewalks, zebra crossings, and cars. This will be our immediate focus for future work, \emph{i.e.}, to advance into global outdoor navigation. 
We also aim to extend our approach to complex scenarios involving multiple individuals. When tested with the scenario with one person, VLM-Social-Nav successfully outputs socially compatible actions, realizing the group of people in the scene. However, when multiple groups are present, simple directions like left or right may not suffice to describe effective robot navigation. 

\section{Conclusion}\label{sec:con}
We propose a novel social navigation approach based on VLMs, focusing on real-time, socially compliant decision-making in human-centric environments. We utilize the perception model to detect important social entities and prompt a VLM to generate guidance for socially compliant behavior. VLM-Social-Nav features a VLM-based scoring that ensures socially appropriate and effective robot actions. This minimizes the dependence on extensive training datasets and eliminates the necessity for explicit rules or {hand-tuned parameters typically associated with imitation learning approaches}. 
By furnishing textual instructions to VLM, we can instruct the robot to adhere to specific navigation rules, such as navigating on the right or left according to cultural norms. 
{However, interpreting social rules and deriving appropriate actions from them based on raw robot perception remains complex. VLMs interpret social situations and determine actions based on these rules, offering a more nuanced approach than simpler rule- or planning-based methods.} 
We demonstrate and evaluate our system in four different real-world social navigation scenarios with a Turtlebot robot.

{One immediate future work is to explore visual prompting methods~\cite{nasiriany2024pivot, liu2024moka} to enhance spatial reasoning in VLMs by marking the images. Another promising future direction is to explore open-source VLMs, such as LLaVA~\cite{liu2024visual}. Their access to lower-level information, such as log probabilities could help detect and address hallucinations. It is also interesting to develop a VLM that generates high-level social navigation instructions through chain-of-thought reasoning, and integrate it into an autonomous navigation system.}

%% file: root_arxiv.bbl
\begin{thebibliography}{10}
\providecommand{\url}[1]{#1}
\csname url@samestyle\endcsname
\providecommand{\newblock}{\relax}
\providecommand{\bibinfo}[2]{#2}
\providecommand{\BIBentrySTDinterwordspacing}{\spaceskip=0pt\relax}
\providecommand{\BIBentryALTinterwordstretchfactor}{4}
\providecommand{\BIBentryALTinterwordspacing}{\spaceskip=\fontdimen2\font plus
\BIBentryALTinterwordstretchfactor\fontdimen3\font minus \fontdimen4\font\relax}
\providecommand{\BIBforeignlanguage}[2]{{%
\expandafter\ifx\csname l@#1\endcsname\relax
\typeout{** WARNING: IEEEtran.bst: No hyphenation pattern has been}%
\typeout{** loaded for the language `#1'. Using the pattern for}%
\typeout{** the default language instead.}%
\else
\language=\csname l@#1\endcsname
\fi
#2}}
\providecommand{\BIBdecl}{\relax}
\BIBdecl

\bibitem{starship}
\BIBentryALTinterwordspacing
S.~Technology. (2024) Starship. [Online]. Available: \url{https://www.starship.xyz/}
\BIBentrySTDinterwordspacing

\bibitem{dilligent}
\BIBentryALTinterwordspacing
D.~Robotics. (2024) Dilligent robotics. [Online]. Available: \url{https://www.diligentrobots.com/}
\BIBentrySTDinterwordspacing

\bibitem{astro}
\BIBentryALTinterwordspacing
Amazon. (2024) Meet astro, a home robot unlike any other. [Online]. Available: \url{https://www.aboutamazon.com/news/devices/meet-astro-a-home-robot-unlike-any-other}
\BIBentrySTDinterwordspacing

\bibitem{mirsky2024conflict}
R.~Mirsky, X.~Xiao, J.~Hart, and P.~Stone, ``Conflict avoidance in social navigation—a survey,'' \emph{ACM Transactions on Human-Robot Interaction}, vol.~13, no.~1, pp. 1--36, 2024.

\bibitem{mavrogiannis2023core}
C.~Mavrogiannis \emph{et~al.}, ``Core challenges of social robot navigation: A survey,'' \emph{ACM Transactions on Human-Robot Interaction}, vol.~12, no.~3, pp. 1--39, 2023.

\bibitem{francis2023principles}
A.~Francis \emph{et~al.}, ``Principles and guidelines for evaluating social robot navigation algorithms,'' \emph{ACM Transactions on Human-Robot Interaction}, 2024.

\bibitem{hirose2023sacson}
N.~Hirose \emph{et~al.}, ``Sacson: Scalable autonomous control for social navigation,'' \emph{IEEE Robotics and Automation Letters}, 2023.

\bibitem{raj2024targeted}
A.~H. Raj \emph{et~al.}, ``Targeted learning: A hybrid approach to social robot navigation,'' in \emph{IEEE International Conference on Robotics and Automation}, 2024.

\bibitem{kretzschmar2016socially}
H.~Kretzschmar \emph{et~al.}, ``Socially compliant mobile robot navigation via inverse reinforcement learning,'' \emph{The International Journal of Robotics Research}, vol.~35, no.~11, pp. 1289--1307, 2016.

\bibitem{thorDataset2019}
A.~Rudenko \emph{et~al.}, ``Th{\"o}r: Human-robot navigation data collection and accurate motion trajectories dataset,'' \emph{IEEE Robotics and Automation Letters}, vol.~5, no.~2, pp. 676--682, 2020.

\bibitem{karnan2022scand}
H.~Karnan \emph{et~al.}, ``Socially compliant navigation dataset (scand): A large-scale dataset of demonstrations for social navigation,'' \emph{IEEE Robotics and Automation Letters}, 2022.

\bibitem{nguyenmusohu}
D.~M. Nguyen, M.~Nazeri, A.~Payandeh, A.~Datar, and X.~Xiao, ``Toward human-like social robot navigation: A large-scale, multi-modal, social human navigation dataset,'' in \emph{IEEE/RSJ International Conference on Intelligent Robots and Systems}, 2023, pp. 7442--7447.

\bibitem{wei2022chain}
J.~Wei \emph{et~al.}, ``Chain-of-thought prompting elicits reasoning in large language models,'' \emph{Advances in Neural Information Processing Systems}, vol.~35, pp. 24\,824--24\,837, 2022.

\bibitem{gpt4v2023}
J.~Achiam, S.~Adler, S.~Agarwal, L.~Ahmad, I.~Akkaya, F.~L. Aleman, D.~Almeida, J.~Altenschmidt, S.~Altman, S.~Anadkat \emph{et~al.}, ``Gpt-4 technical report,'' \emph{arXiv preprint arXiv:2303.08774}, 2023.

\bibitem{wen2024road}
L.~Wen \emph{et~al.}, ``On the road with gpt-4v (ision): Explorations of utilizing visual-language model as autonomous driving agent,'' in \emph{ICLR 2024 Workshop on Large Language Model (LLM) Agents}, 2024.

\bibitem{shah2022lmnav}
D.~Shah \emph{et~al.}, ``Lm-nav: Robotic navigation with large pre-trained models of language, vision, and action,'' 2022.

\bibitem{YOLO2023}
J.~Redmon, S.~Divvala, R.~Girshick, and A.~Farhadi, ``You only look once: Unified, real-time object detection,'' in \emph{Proc. of the IEEE Conf. on Computer Vision and Pattern Recognition}, 2016, pp. 779--788.

\bibitem{fox1997dynamic}
D.~Fox, W.~Burgard, and S.~Thrun, ``The dynamic window approach to collision avoidance,'' \emph{IEEE Robotics \& Automation Magazine}, vol.~4, no.~1, pp. 23--33, 1997.

\bibitem{pomerleau1988alvinn}
D.~A. Pomerleau, ``Alvinn: An autonomous land vehicle in a neural network,'' \emph{Advances in neural information processing systems}, vol.~1, 1988.

\bibitem{song2024socially}
D.~Song, J.~Liang, A.~Payandeh, X.~Xiao, and D.~Manocha, ``Vlm-social-nav: Socially aware robot navigation through scoring using vision-language models,'' \emph{arXiv preprint arXiv:2404.00210}, 2024.

\bibitem{liang2021vo}
J.~Liang, Y.-L. Qiao, T.~Guan, and D.~Manocha, ``Of-vo: Efficient navigation among pedestrians using commodity sensors,'' \emph{IEEE Robotics and Automation Letters}, vol.~6, no.~4, pp. 6148--6155, 2021.

\bibitem{liang2021crowd}
J.~Liang, U.~Patel, A.~J. Sathyamoorthy, and D.~Manocha, ``Crowd-steer: Realtime smooth and collision-free robot navigation in densely crowded scenarios trained using high-fidelity simulation,'' in \emph{Proc. of the Twenty-Ninth International Conference on International Joint Conferences on Artificial Intelligence}, 2021, pp. 4221--4228.

\bibitem{best2014densesense}
A.~Best \emph{et~al.}, ``Densesense: Interactive crowd simulation using density-dependent filters.'' in \emph{Symposium on Computer Animation}, 2014, pp. 97--102.

\bibitem{gopalakrishnan2017prvo}
B.~Gopalakrishnan \emph{et~al.}, ``Prvo: Probabilistic reciprocal velocity obstacle for multi robot navigation under uncertainty,'' in \emph{IEEE/RSJ International Conference on Intelligent Robots and Systems}, 2017, pp. 1089--1096.

\bibitem{sun2021motion}
H.~Sun, W.~Zhang, R.~Yu, and Y.~Zhang, ``Motion planning for mobile robots—focusing on deep reinforcement learning: A systematic review,'' \emph{IEEE Access}, vol.~9, pp. 69\,061--69\,081, 2021.

\bibitem{aradi2020survey}
S.~Aradi, ``Survey of deep reinforcement learning for motion planning of autonomous vehicles,'' \emph{IEEE Transactions on Intelligent Transportation Systems}, vol.~23, no.~2, pp. 740--759, 2020.

\bibitem{7539621}
P.~Liu, D.~F. Glas, T.~Kanda, and H.~Ishiguro, ``Data-driven hri: Learning social behaviors by example from human–human interaction,'' \emph{IEEE Transactions on Robotics}, vol.~32, no.~4, pp. 988--1008, 2016.

\bibitem{li2018role}
M.~Li, R.~Jiang, S.~S. Ge, and T.~H. Lee, ``Role playing learning for socially concomitant mobile robot navigation,'' \emph{CAAI Transactions on Intelligence Technology}, vol.~3, no.~1, pp. 49--58, 2018.

\bibitem{nazeri2024vanp}
M.~Nazeri, J.~Wang, A.~Payandeh, and X.~Xiao, ``Vanp: Learning where to see for navigation with self-supervised vision-action pre-training,'' 2024.

\bibitem{bommasani2022opportunities}
R.~o. Bommasani, ``On the opportunities and risks of foundation models,'' \emph{arXiv preprint arXiv:2108.07258}, 2022.

\bibitem{Ahn2022}
M.~Ahn \emph{et~al.}, ``Do as i can, not as i say: {{Grounding}} language in robotic affordances,'' 2022.

\bibitem{mao2023gptdriver}
J.~Mao, Y.~Qian, J.~Ye, H.~Zhao, and Y.~Wang, ``Gpt-driver: Learning to drive with gpt,'' 2023.

\bibitem{yu2023l3mvn}
B.~Yu, H.~Kasaei, and M.~Cao, ``L3mvn: Leveraging large language models for visual target navigation,'' in \emph{IEEE/RSJ International Conference on Intelligent Robots and Systems}, 2023, pp. 3554--3560.

\bibitem{li2024driving}
B.~Li, Y.~Wang, J.~Mao, B.~Ivanovic, S.~Veer, K.~Leung, and M.~Pavone, ``Driving everywhere with large language model policy adaptation,'' 2024.

\bibitem{radford2021learning}
A.~Radford \emph{et~al.}, ``Learning transferable visual models from natural language supervision,'' in \emph{International conference on machine learning}.\hskip 1em plus 0.5em minus 0.4em\relax PMLR, 2021, pp. 8748--8763.

\bibitem{duan2024manipulate}
J.~Duan, W.~Yuan, W.~Pumacay, Y.~R. Wang, K.~Ehsani, D.~Fox, and R.~Krishna, ``Manipulate-anything: Automating real-world robots using vision-language models,'' \emph{arXiv preprint arXiv:2406.18915}, 2024.

\bibitem{min2022rethinking}
S.~Min \emph{et~al.}, ``Rethinking the role of demonstrations: What makes in-context learning work?'' \emph{arXiv preprint arXiv:2202.12837}, 2022.

\bibitem{kruse2013human}
T.~Kruse, A.~K. Pandey, R.~Alami, and A.~Kirsch, ``Human-aware robot navigation: A survey,'' \emph{Robotics and Autonomous Systems}, vol.~61, no.~12, pp. 1726--1743, 2013.

\bibitem{tsoi2024robot}
N.~Tsoi, J.~Romero, and M.~V{\'a}zquez, ``How do robot experts measure the success of social robot navigation?'' in \emph{Companion of the 2024 ACM/IEEE International Conference on Human-Robot Interaction}, 2024, pp. 1063--1066.

\bibitem{pirk2022protocol}
S.~Pirk, E.~Lee, X.~Xiao, L.~Takayama, A.~Francis, and A.~Toshev, ``A protocol for validating social navigation policies,'' \emph{arXiv preprint arXiv:2204.05443}, 2022.

\bibitem{nasiriany2024pivot}
S.~Nasiriany \emph{et~al.}, ``Pivot: Iterative visual prompting elicits actionable knowledge for vlms,'' \emph{arXiv preprint arXiv:2402.07872}, 2024.

\bibitem{liu2024moka}
F.~Liu, K.~Fang, P.~Abbeel, and S.~Levine, ``Moka: Open-vocabulary robotic manipulation through mark-based visual prompting,'' in \emph{First Workshop on Vision-Language Models for Navigation and Manipulation at ICRA 2024}, 2024.

\bibitem{liu2024visual}
H.~Liu, C.~Li, Q.~Wu, and Y.~J. Lee, ``Visual instruction tuning,'' \emph{Advances in neural information processing systems}, vol.~36, 2024.

\end{thebibliography}
